\documentclass{article}

\PassOptionsToPackage{numbers}{natbib}

 \usepackage[preprint]{neurips_2026}

\usepackage[utf8]{inputenc} 
\usepackage[T1]{fontenc}    
\usepackage{hyperref}       
\usepackage{url}            
\usepackage{booktabs}       
\usepackage{amsfonts}       
\usepackage{nicefrac}       
\usepackage{microtype}      
\usepackage{xcolor}   
\usepackage{graphicx} 

\usepackage{placeins}
\usepackage[symbol]{footmisc}
\usepackage{float}

\usepackage{multirow}
 \usepackage{amsmath}
 \usepackage{enumitem}

\usepackage{multirow}
\usepackage{colortbl}
\usepackage{pifont}
\usepackage{fontawesome5} 
 
\usepackage{marvosym}

\definecolor{cPrompt}{HTML}{A8487E}      
\definecolor{cOffline}{HTML}{005A42}     
\definecolor{cOnline}{HTML}{8C3D00}      
\definecolor{cContext}{HTML}{00557F}     
\definecolor{cFullBg}{HTML}{EEECE6}      
\definecolor{cSparseBg}{HTML}{D6EAF7}    
\definecolor{cSparseFg}{HTML}{00405C}
\definecolor{cCheck}{HTML}{009E73}       

\newcommand{\full}{\colorbox{cFullBg}{\small Full}}
\newcommand{\sparse}{\colorbox{cSparseBg}{\textcolor{cSparseFg}{\small Sparse}}}
\newcommand{\cmark}{\textcolor{cCheck}{\ding{51}}}
\newcommand{\xmark}{\textcolor{black!40}{\ding{55}}}

\title{When Does Continual Learning Require Learning}

\author{%
  Anne Harrington\textsuperscript{1,*} \enspace
  Nayan Saxena\textsuperscript{2} \enspace
  Michael Murphy\textsuperscript{1} \enspace 
  Anastasia Borovykh\textsuperscript{3}\\
  \textbf{Zeyu Yun}\textsuperscript{1} \quad \quad
  \textbf{Sridhar Kamath}\textsuperscript{2} \quad \quad
  \textbf{Ara Eindra Kyi}\textsuperscript{2} \\
  \textbf{Trevor Darrell}\textsuperscript{1} \quad \quad
  \textbf{Jitendra Malik}\textsuperscript{1} \quad \quad
  \textbf{Yutong Bai}\textsuperscript{1,*} \\ \\
  \textsuperscript{1}UC Berkeley \qquad
  \textsuperscript{2}Independent \qquad
  \textsuperscript{3}Capital Fund Management\\ 
  \texttt{\{anneko, yutongbai\}@berkeley.edu}
}

\begin{document}

\maketitle

\begin{abstract}

As large language models (LLMs) become increasingly capable, the next question is how can we enable models to continually learn? Today, the field largely frames this as a problem of context management and mitigating forgetting. We argue this framing is incomplete: continual learning is fundamentally about increasing model competence as the world changes. We disentangle this change along two axes — space, where the model encounters new domains, and time, where the underlying data drifts under a fixed task. This framing lets us study continual learning under realistic conditions: new domains arrive over time, facts drift past their training cutoff, and agentic interactions accumulate state across episodes. To evaluate methods under this setting, we recast widely used LLM benchmarks as sequential problems and introduce a single mechanism-agnostic protocol that compares prompt-based methods (GEPA, ACE), supervised learning (SFT, SDFT), reinforcement learning (GRPO, SDPO), and context compression (Cartridges, In-place TTT). Prompt-based methods fit each new stage quickly but degrade on future tasks. Distillation-based methods accumulate knowledge stably but struggle to update outdated facts. Context compression improves efficiency without substantially improving the ability to learn new tasks. Online reinforcement learning adapts most effectively to knowledge updates but remains sensitive to noisy reward signals.  
Overall, our results suggest that continual learning is not a single capability: different patterns of environmental change require fundamentally different update behaviors, determining when adaptation must be learned inside model weights and when it can be achieved through external scaffolding. We hope that understanding where each method succeeds and fails will guide the design of stronger continual learning systems.$^{\dagger}$

\end{abstract}


\section{Introduction}
\footnotetext[1]{Equal Contribution. \quad $^{\dagger}$Code: \url{https://github.com/anneharrington/studying-cl}}

\begin{figure}
    \centering
    \makebox[\textwidth][c]{%
      \includegraphics[width=1.2\linewidth]{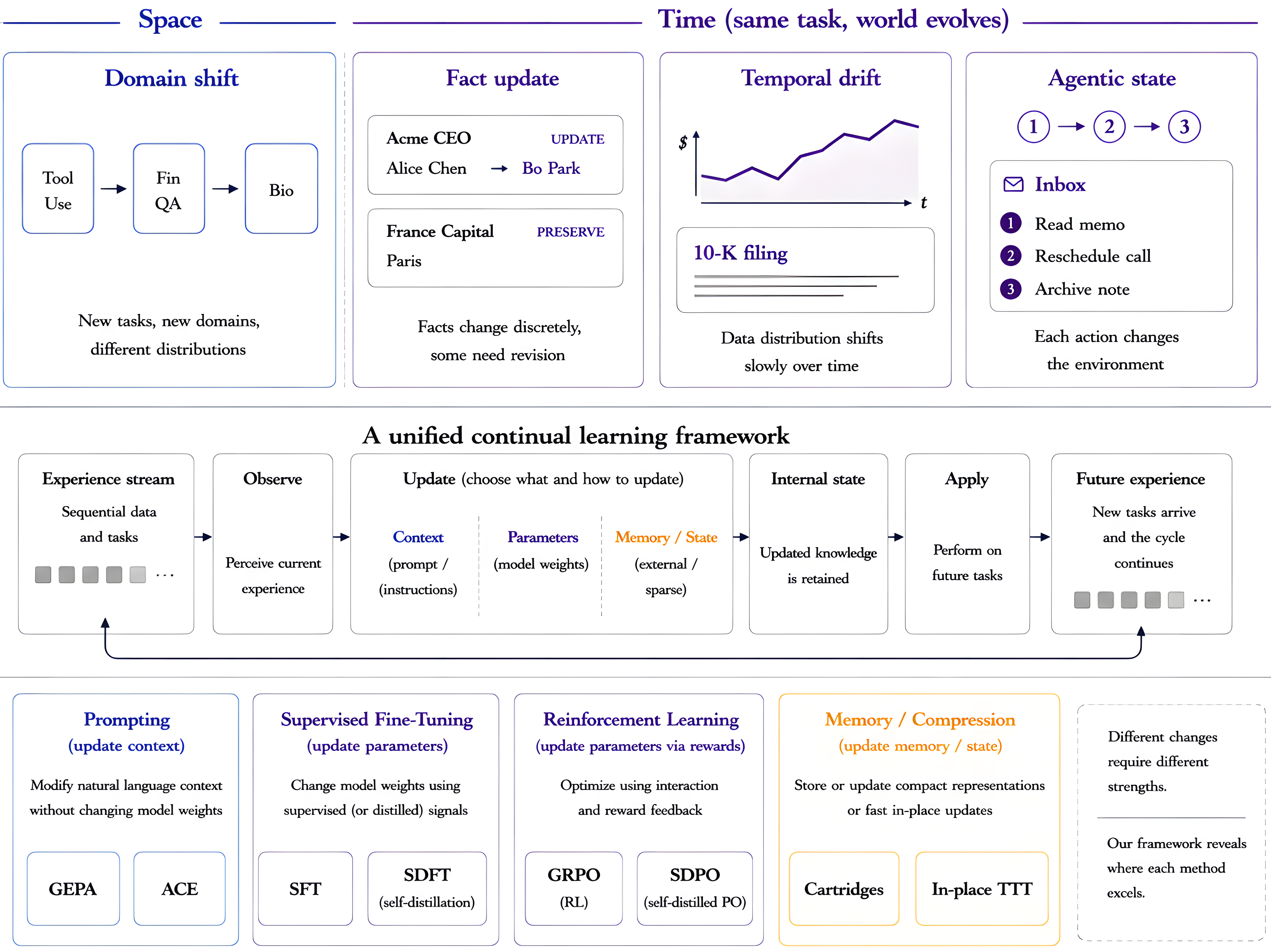}%
    }
    \caption{\textbf{A unified continual learning framework spanning axes of change and update mechanisms.} We identify four axes along which tasks evolve: domain shift, fact update, temporal drift, and agentic state. All methods are cast within a common framework where an experience stream triggers updates to one of three internal components: context, parameters, or memory/state. Existing methods are unified accordingly: prompting methods (GEPA, ACE) modify context; fine-tuning methods (SFT, SDFT) and reinforcement learning methods (GRPO, SDPO) update parameters; and memory/compression methods (Cartridges, In-place TTT) maintain external state. }
    \label{fig:fig1}
\end{figure}

Traditionally, continual learning has been defined as mitigating catastrophic forgetting \cite{mccloskey1989catastrophic}:
\emph{how do we learn task $B$ without forgetting task $A$?} One approach to this problem is to augment or compose model parameters as proposed in superposition \cite{cheung2019superposition}, adversarial networks \cite{ebrahimi2020adversarial} and side networks \cite{zhang2020side}. However, this setup is not realistic in the current paradigm of large language models (LLMs), which often rely on large scale pre-, mid-, and post-training to acquire general capability and after deployment need to be adapted to a \emph{broad} range of potential use cases. At the same time, standard LLM evaluations focus on measuring capabilities on static tasks such as general knowledge \cite{wang2024mmlu}, domain specific abilities \cite{hendrycks2021measuring, jimenez2023swe, ouyang2025kernelbench, grady2026kellybench}, and challenging frontier problems \cite{glazer2024frontiermath, mang2025frontiercs}; these however fail to measure the ability to become more competent under conditions that evolve through time -- which is critical for practical deployment. 

Without a unified framework, the community  approached continual learning
as a memory problem solved with retrieval-augmented generation \cite{lewis2020retrieval}, in-context learning \cite{brown2020language}, smart context management \citep{li2023compressing, zhang2025recursive} and harnesses \cite{zhang2025agentic, ye2026meta, lee2026meta}; as a steering problem solved with prompt optimization \cite{novikov2025alphaevolve, openevolve, agrawal2025gepa}; as a reasoning problem solved with reinforcement learning and distillation \cite{lu2025onpolicydistillation, zhao2026self, sdpo, shenfeld2026self}; and as a compression problem solved with architectural changes \cite{eyuboglu2025cartridges, lin2025continual, tandon2025end, feng2026place}. 
Each of these takes a different bet on what continual learning actually requires, but the field still lacks unified insights across these methods to direct progress.

In our work, we provide a unified framework to define and measure continual learning in modern LLMs. We define continual learning as the problem of increasing competence as the world changes. We consider change along two axes: \emph{space} and \emph{time}, as depicted in Figure \ref{fig:fig1}. Space corresponds to the classical notion of domain or task shift, where the model is asked to acquire a new skill on a different distribution without losing the previous one. Time corresponds to a different and, more common form of change in real deployed systems: the task stays the same, but the underlying world drifts. Time itself can be further split into two sub-regimes that the literature usually conflates. Slow trends (e.g.\ yearly financial filings) ask the model to stay aligned; the best strategy here is to keep extracting whatever structure survives across years and ignore year-specific noise. Discrete fact changes (e.g.\ Wikipedia revisions) ask the opposite -- the right move is to actually rewrite the affected belief while leaving the rest of the model's knowledge untouched. A third sub-regime is agentic accumulation: the task family itself stays fixed (e.g., actions inside a web app), but the environment’s state drifts as a direct consequence of the model’s own prior actions, so that what one step leaves behind becomes part of the task the model faces at the next step.
This second sub-regime is the LLM analogue of the frame problem \cite{mccarthy1981some}: when the world updates, a learner must decide which of its beliefs are now outdated and which can remain fixed.

 To measure continual learning under our definition, we cast widely used LLM evaluations as problems that unfold sequentially under a single mechanism-agnostic protocol, so prompt, weight, and architectural updates can be compared on equal ground. On a common backbone (Qwen3-8B) \cite{yang2025qwen3}, we evaluate eight methods across four families: prompt optimization (GEPA \cite{agrawal2025gepa}, ACE \cite{zhang2025agentic}), offline supervised updates (SFT, SDFT \cite{shenfeld2026self}), online reinforcement learning (GRPO \cite{shao2024grpo}, SDPO \cite{sdpo}), and context-compression (Cartridges \cite{eyuboglu2025cartridges}, In-place TTT \cite{feng2026place}).

Interestingly, we observe a consistent set of tradeoffs across methods: (i) Prompting-based approaches such as GEPA and ACE fit quickly and heavily to past trajectories, achieving strong backward accuracy, but encounter sharp degradation on future tasks. (ii) Distillation-based methods such as SDFT and SDPO accumulate knowledge more stably over time, yet struggle to rapidly incorporate new information or update outdated knowledge. (iii) Context compression approaches, including Cartridges and In-place TTT, improve efficiency and memory management but do not substantially improve the ability to learn new tasks. (iv) In contrast, online reinforcement learning methods such as GRPO adapt more effectively to knowledge updating, though they remain highly sensitive to noisy or unstable reward signals. (v) On the agentic axis, where the stage ordering is generated by the agent’s own actions rather than imposed by us, both a prompt-based playbook (ACE) and weight-based fine-tuning (SFT) compound experience and beat their zero-shot baselines at every chain length, though absolute success still decays as the chain grows.

All together, we make the following contributions:
\begin{itemize}[leftmargin=.5em]
    \item We formalize continual learning in LLMs as the problem of increasing competence over time. Through our protocol, we provide a unified perspective that allows us to consider both parametric and non-parametric update methods for continual learning. We adapt these methods to the continually learning setting and align their evaluation metrics. 
    \item We introduce a broad suite of sequential LLM tasks to evaluate continual learning. These including domain adaptation, agentic tasks, financial analysis, and temporally-dependent knowledge updates.
    \item We find that prompt-based methods alone are insufficient across most regimes, calling for actual learning. Different learning methods each have their own strengths depending on how the task and data shift over time.
\end{itemize}

\section{Related Works}





\noindent \textbf{Continual Learning.}
Traditionally, the field has considered continual learning as preventing the problem of catastrophic forgetting~\citep{mccloskey1989catastrophic,french1999catastrophic}. In this context, the problem is that sequential parameter updates can overwrite knowledge needed for previous tasks; this led to a focus on designing methods that reduce task interference~\cite{javed2019meta,golkar2019continual,ebrahimi2020adversarial,zhang2020side,cheung2019superposition}. From another angle, continual learning has been positioned as maintaining plasticity~\cite{dohare2024loss}, where models must not only remember old tasks, but also have the ability to learn new ones~\cite{lyle2022understanding,nikishin2022primacy}. 


\textbf{Non-parametric updates.}
At deployment, we can adapt LLMs without weight changes by changing the inference-time environment around a fixed model. One family of methods changes the information available to the model, for example through retrieval-augmented generation (RAG) \cite{lewis2020retrieval}, external memory, reusable long-context representations~\cite{eyuboglu2025cartridges}, context compression~\cite{li2023compressing}, or recursive access to large external contexts~\cite{zhang2025recursive}. A closely related family changes the behavioral scaffold through which that information is used, including demonstrations~\cite{brown2020language}, prompt instructions and agent contexts~\cite{agrawal2025gepa, zhang2025agentic, yuksekgonul2024textgrad} and multi-stage LM programs~\cite{opsahl2024optimizing, khattab2023dspy,li2026combee}. More broadly, agentic systems can adapt by modifying their surrounding harnesses~\cite{ye2026meta, lee2026meta}, learn component-level reward specifications~\cite{wu2025optimas} or improve through external programs and agent code modifications~\cite{cemri2026adaevolve, novikov2025alphaevolve, zhang2025darwin, openevolve}. 


\textbf{Gradient-based (parametric) updates.}
Another adaption approach is to update weights or activations. At one extreme, full fine-tuning modifies all parameters of the model; while effective, this regime is often computationally expensive and unrealistic for frequent or online adaptation. Activation editing \cite{ilharco2022editing} uses task vectors applied to all parameters. More commonly, work studies \emph{parameter-efficient} methods, such as LoRA \cite{hu2022lora} and adapter-based tuning \cite{lin2025continual,lopez2017gradient}, which restrict updates to small subsets of parameters. An important but underexplored regime involves severely constrained updates—very few gradient steps and tightly limited compute—closer to the conditions under which continual adaptation would need to occur in deployed systems. Continual learning methods differ not only in where they store updates, but also in what learning signal drives the update. OPSD~\cite{zhao2026self, lu2025onpolicydistillation}, SDFT~\cite{shenfeld2026self} and SDPO \cite{sdpo} learn from self-distillation, GRPO \cite{shao2024grpo} and h1 \cite{motwani2025h1} optimize end-of-task rewards, LoRD \cite{liu2025low} uses teacher–student distillation, sparse memory finetuning \cite{lin2025continual} localizes the next-token objective to sparsely accessed memory slots, test-time training \cite{feng2026place} performs input-conditioned weight updates during inference. 




\section{Method}
\label{sec:framework}



\subsection{A Protocol for Continual Learning}

\noindent \textbf{Setup.}
Typically, continual-learning evaluations are tightly coupled to a particular method: weight-update methods use task accuracy with a fixed
fine-tuning budget; prompt-optimization often uses few-shot generalization
with a fixed token budget; agentic frameworks rely on end-to-end pass rate.
As a
result, prompt, weight, and architectural updates are rarely compared fairly against one another. To gather broader insights about what methods are necessary for continual learning, we define a simple protocol that is compatible with any update operator. To do so, we consider all forms of updates (prompting, full weight, sparse weight/activation) as potential strategies, and group methods into families based on which aspect of the model they change during optimization (Tab.~\ref{tab:methods}). Formally:
 
A learner sees stages $\mathcal{T}_1, \dots, \mathcal{T}_K$ in fixed
order. Each $\mathcal{T}_k$ specifies a training distribution
$\mathcal{D}_k^{\text{tr}}$ and an evaluation set
$\mathcal{D}_k^{\text{eval}}$. Between stages the learner applies an
update operator
\[
\theta_k \;=\; \mathcal{U}_k\!\left(\theta_{k-1},\, \mathcal{D}_k^{\text{tr}}\right),
\qquad \theta_0 = \theta_{\text{base}}.
\]
$\mathcal{U}_k$ is unrestricted: it may modify model weights, edit a system
prompt, write to or read from external memory, attach an adapter, or
perform no update. A method is fully specified by its family of
$\mathcal{U}_k$ and a per-stage compute budget $C$ (held constant across
methods within a benchmark).
 
\noindent \textbf{Forgetting matrix.}
After each stage we evaluate on every stage's eval set, producing
\[
R_{i,j} \;=\; \mathrm{acc}\!\left(\theta_i,\, \mathcal{D}_j^{\text{eval}}\right),
\qquad i \in \{0, \dots, K\},\; j \in \{1, \dots, K\},
\]
with row $0$ the base model before any update. We report two transfer
quantities:
\begin{align*}
\textsc{BWT} &= \tfrac{1}{K-1}\textstyle\sum_{k=1}^{K-1} \big(R_{K,k} - R_{k,k}\big),
&
\textsc{FWT} &= \tfrac{1}{K-1}\textstyle\sum_{k=2}^{K} \big(R_{k-1,k} - R_{0,k}\big).
\end{align*}
\textsc{BWT} \citep{lopez2017gradient} measures whether later stages
improve or damage earlier ones (negative \textsc{BWT} is catastrophic
forgetting). \textsc{FWT} measures whether prior stages prepare the model
for stages it has not yet trained on, and is the sharper signal: it tests
acquisition of structure that transfers, rather than fitting the current
stage at the cost of the rest.

\subsection{Update Methods}
\label{sec:benchmarks}

We evaluate eight methods across four families: prompt
optimization, supervised weight updates, reinforcement learning, and
per-stage architectural updates. All use Qwen3-8B (non-thinking)
\cite{yang2025qwen3} as $\theta_0$ and run under the protocol of
\S\ref{sec:framework}. Table~\ref{tab:methods} states what each method
carries across the stage boundary; the per-family paragraphs that
follow specify each $\mathcal{U}_k$ and its sequential adaptation.



\begin{table}[t]

\centering
\caption{The eight methods, organized by what crosses the
stage boundary. ``Sequential adaptation'' is the modification that
lifts a single-stage method to the staged protocol of
\S\ref{sec:framework}. Prompting refers to methods that update in natural language. Offline corresponds to supervised learning whereas online refers to reinforcement learning. Lastly compression notes methods that compress context in activation or weight space. }
\small
\setlength{\tabcolsep}{4pt}
\renewcommand{\arraystretch}{1.15}
\begin{tabular}{@{}llllcc@{}}
\toprule
\textbf{Family} & \textbf{Method} & \textbf{Carried} & \textbf{Sequential adaptation} & \textbf{Distill.} & \textbf{Weights} \\
\midrule
\textbf{Prompting }  & GEPA \citep{agrawal2025gepa}              & evolved prompt       & init from $k{-}1$ prompt        & \xmark & ---     \\
   & ACE \citep{zhang2025agentic}                  & markdown playbook    & init from $k{-}1$ playbook      & \xmark & ---     \\
\midrule
\textbf{Offline }   & SFT                                       & full params          & resume from $\theta_{k-1}$      & \xmark & \full   \\
    & SDFT \citep{shenfeld2026self}             & params + teacher     & $\theta_{k-1}$ as teacher       & \cmark & \full   \\
\midrule
\textbf{Online  }     & GRPO \citep{shao2024grpo}                 & full params          & resume from $\theta_{k-1}$      & \xmark & \full   \\
 & SDPO \citep{sdpo}                         & params + pref.\ pool & $\theta_{k-1}$ generates pairs  & \cmark & \full   \\
\midrule
\textbf{Context} & Cartridges \citep{eyuboglu2025cartridges} & backbone + adapter   & new adapter per stage           & \cmark & \sparse \\
\textbf{Compression} & In-place TTT \citep{feng2026place}   & per-input update     & reset between inputs            & \xmark & \sparse \\
\bottomrule
\end{tabular}
\vspace{1em}


\label{tab:methods}
\vspace{-2em}
\end{table}
%

\noindent \textbf{Prompt optimization (GEPA, ACE)}
Both methods leave model weights unchanged. GEPA \citep{agrawal2025gepa}
optimizes the system prompt through a reflect-and-mutate loop scored on
a validation minibatch; the best-scoring mutations propagate forward. ACE
\citep{zhang2025agentic} maintains a markdown playbook edited incrementally
through explicit add/keep/drop operations, which the authors show resists
brevity-bias and context collapse. In our protocol, $\mathcal{U}_k$
initializes optimization for stage $k$ from the optimized state at the
end of stage $k{-}1$ (the evolved prompt for GEPA, the playbook for ACE),
prepended to the default instruction for $\mathcal{T}_k$. No replay of
prior data is performed. Apparent forward transfer is therefore
attributable to the carried prompt or playbook, not to revisiting
training examples.
 
\noindent \textbf{Supervised / Offline Learning (SFT, SDFT)}
\textbf{SFT} fine-tunes all parameters on $\mathcal{D}_k^{\text{tr}}$,
resuming from $\theta_{k-1}$. \textbf{SDFT} \citep{shenfeld2026self} is
self-distillation under a forward-KL loss against soft targets from a
teacher; in the original work the teacher is the model itself on
demonstration-conditioned inputs. Our sequential adaptation: at stage
$k$, the teacher is $\theta_{k-1}$, so the student learns the new stage
under a regularizer pulling it toward its own previous-stage
distribution. This is the smallest modification that lifts SDFT to a
continual-learning operator and is the only modification we make. SFT
and SDFT share a per-stage update budget within $C$ and differ only in
the loss applied at each step.
 
\noindent \textbf{Reinforcement / Online Learning (GRPO, SDPO)}
\textbf{GRPO} \citep{shao2024grpo} samples rollouts in groups, computes
within-group advantages from a verifiable reward, and updates the policy
under a KL constraint. For benchmarks without a verifiable reward
(10-K, TempWiki) we use a binary correctness signal from the
ground-truth label. \textbf{SDPO} \citep{sdpo} is sequential DPO. Our
adaptation: at stage $k$, $\theta_{k-1}$ generates both the chosen and
rejected continuation pools, so the preference signal is implicitly
relative to the model's accumulated prior rather than to a fresh
reference. Both methods consume $C$ as policy-update steps.
 
\noindent \textbf{Architectural / Context Compression (Cartridges, In-place TTT)}
These methods change the model's update interface rather than the loss
applied to its parameters. \textbf{Cartridges}
\citep{eyuboglu2025cartridges} freezes the backbone and learns a small
per-stage component, distilled from a teacher trace and stitched in at
inference. The backbone is $\theta_0$ for every stage; each stage's
update is structurally local. \textbf{In-place TTT}
\citep{feng2026place} performs an input-conditioned weight update
at inference using a self-supervised auxiliary loss, reset between
inputs. We include In-place TTT as a locality reference: its update is
per-input rather than per-stage, which lets us isolate the locality
axis from the accumulation axis in the experiment.

\section{Experiments}

For each task evaluation, we apply our general continual learning protocol and vary only how the world changes between optimization stages.
We work with the two axes of change: \textbf{space}, where the model
moves across substantively different domains, and \textbf{time}, where
the model stays in one domain but the underlying data drifts across
stages. Within these axes we instantiate four settings:
\emph{independent domains} (space), where each stage is a different task; \emph{selective fact change}
(time), where some labels change between stages while others remain
stable;
\emph{continuous temporal drift} (time), where the task is fixed but the
data slides smoothly across stages; and  and \emph{agentic accumulation} (time), where the task family is fixed but the environment’s state drifts as a direct consequence of the agent’s own prior actions rather than an external clock or editor. More details on task design choices can be found in Appendix~\ref{appendix:tasks}.

\subsection{Domain Transfer}
\label{sec:domain}



When a model is exposed to a new domain, it will need to acquire new skills or capabilities without catastrophically forgetting. This is our evaluation setting and the classic test of continual learning and covers the axis of space. 

\noindent \textbf{Setup.} Three unrelated supervised tasks are presented in a fixed order — ToolUse~\cite{tang2023toolalpaca}, FinQA~\cite{chen2021finqa}, SciKE-Bio~\cite{feng2024sciknoweval} with each phase training a Qwen-3-8B on a 500-example subsample of one task's training set and validating against all three tasks at every step. The subset of ToolUse and SciKE-Bio are taken from \cite{shenfeld2026self}. The task chain produces a forgetting matrix that answers "after seeing tasks $1...i$, how well does the model still do task $j$?" The three tasks deliberately share no surface-level structure, so a method that nominally improves task acquisition can still fail by drifting the model's output format or by overwriting earlier formats once it learns a later one. For continual training, the best performing checkpoint on the current task is carried over to the next one. Further details about the experimental setup including relevant hyperparameters are included in Appendix \ref{appendix:domain}.9

\begin{figure}[!h]
    \centering
    \includegraphics[width=1\linewidth]{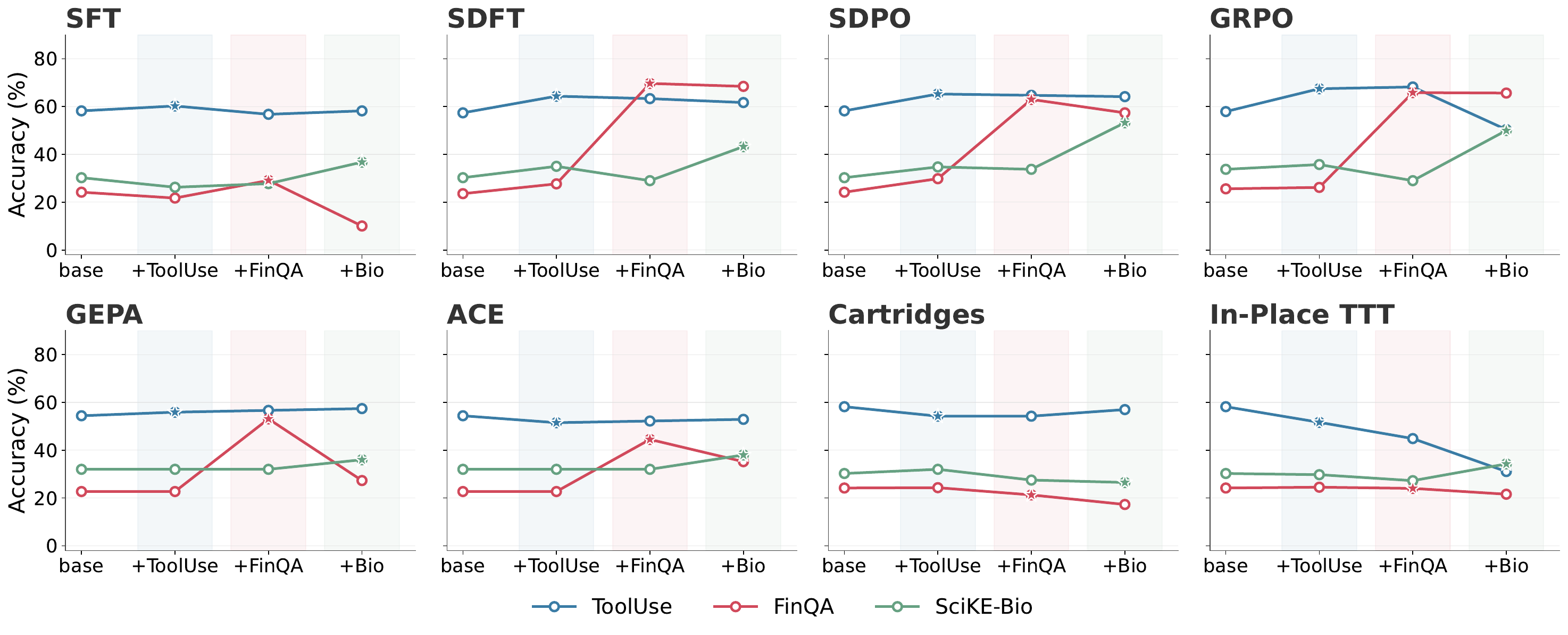}
    \vspace{-2em}
    \caption{Per-method accuracy across the three-stage domain chain
(ToolUse $\to$ FinQA $\to$ SciKE-Bio) for Qwen3-8B (non-thinking). Each panel shows one method; the three colored curves are
held-out accuracy on each of the three eval sets across all training
steps, with the corresponding panels marking stage boundaries. Diagonal acquisition, the curve corresponding to the currently-trained task, and
the other two curves showing off-diagonal forgetting, are visible side by side.
}
    \label{fig:domaintransfer}
\end{figure}








\noindent \textbf{Prompt and harness optimization surfaces a speed-vs-stability
trade-off.} We observe that GEPA fits very quickly to some training stages such as FinQA, but has the most severe catastrophic forgetting when moving on to the next stage. In Fig.~\ref{fig:domaintransfer} we observe that despite an almost 40\% increase in performance during the FinQA training stage, after the training on biology accuracy drops almost entire back down the baseline accuracy before any training. While ACE appears to be slight more stable than GEPA, it similarly improves and then immediately degrades at this same phase transition.

\noindent \textbf{Distillation-based methods manage to accumulate knowledge.} 
We observe that SDFT and SDPO both finish the sequential training
with FinQA and Bio above their zero-shot baselines while keeping ToolUse
near 55--60\%; SDFT is the only method in our table whose three final-stage
scores are all at or above their respective zero-shot levels. 


\noindent \textbf{Context compression does not help learn new tasks.} Cartridges and In-place TTT are less sensitive to than prompt
methods to domain shift, but at the same time they do not show steady improvement across all tasks over training phases like SDFT. Performance is largely flat or even simply degrades per task at each progressive training phase. These methods never perform a full weight update, and thus appear to struggle to accumulate new capabilities under the domain shift task.

\subsection{Catastrophic Memorizing: Update Truly New Knowledge}

In the previous experiment, we tested methods on their ability to stay aligned and accumulate knowledge across time. Here, we ask the opposite: when presented with discrete fact changes, can a model rewrite the affected belief while leaving the rest of its knowledge untouched? Avoiding catastrophic forgetting is necessary, but so is knowing when and how to update. The failure to do so is what we call 'catastrophic memorizing'.





\noindent \textbf{Data.} 
We construct the chain from four monthly Wikipedia mirrors covering the
post-pretraining-cutoff window: 2025-11-20, 2025-12-01, 2026-01-01, and
2026-02-01. For each adjacent pair of snapshots we extract a
\emph{drift set} of (subject, relation, object) triples whose object
value changed between the snapshots, treating these as new world-knowledge
facts the model is being asked to acquire. We also extract a
\emph{stable set} of (subject, relation) pairs whose object did not
change across the entire window, and hold this set out across all
phases as a forgetting probe: a method that improves drift accuracy
by overwriting the model's general world knowledge will pay for it on
the stable probe. 

\noindent \textbf{Setup.} To test for small volume adversarial updates evolving over time on Qwen-3-8B, Wikipedia is mirrored at four monthly snapshots, where every method updates from the same Wikipedia diff snapshot, in its method-appropriate format. Weight-update and RL methods (SFT, SDFT, SDPO, GRPO) consume the 500 drift triples directly as Q/A pairs. Cartridges and In-place TTT
consume the corresponding Wikipedia article bodies pulled from
per-snapshot -- approximately 528 articles per slice. Further details about the experimental setup including relevant hyperparameters are included in Appendix \ref{appendix:wiki}.  A filtered, continuously-scored variant of this benchmark (TempWiki-Easy), which isolates the update behavior from noisy diff categories and binary scoring, is analyzed in Appendix~\ref{appendix:wiki_easy}.

\begin{figure}[!h]
    \centering
    \includegraphics[width=\linewidth]{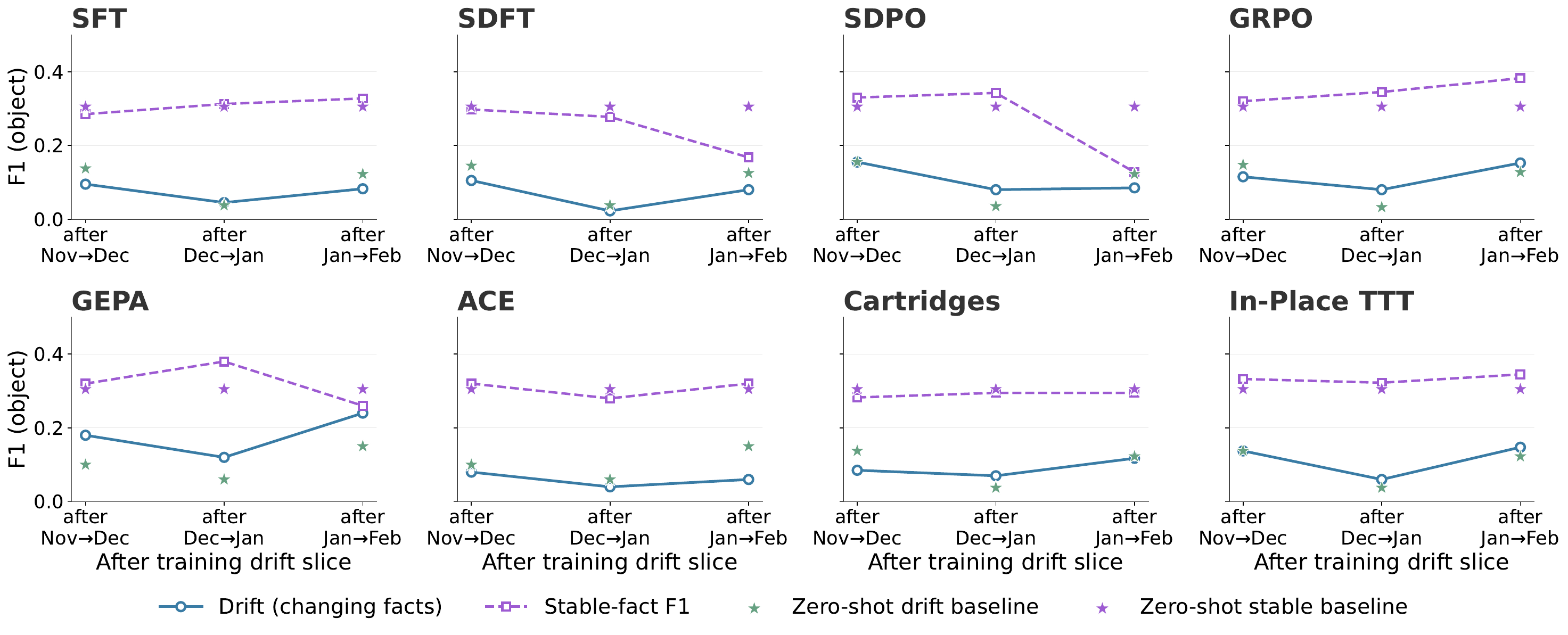}
    \vspace{-2em}
    \caption{F1 score  across
the three TempWiki slices for each method (Qwen3-8B, no thinking).
A method that genuinely learns the new facts should improve drift F1
while preserving stable F1; methods that conflate the two appear as
diverging curves. Cartridges and In-place TTT preserve stable F1 most
cleanly; SDFT and SDPO degrade stable F1 sharply; GEPA and ACE achieve
the largest drift gains but pay heavily on stable. GRPO is the only
weight-update method whose stable F1 trajectory does not move in the
opposite direction of its drift trajectory}
\vspace{-1em}
    \label{fig:tempwiki}
\end{figure}


\noindent \textbf{Stability-anchored methods fail to learn how to change.}  Interestingly, SDFT and SDPO, which perform well on accumulating knowledge in noisy signal, struggle to learn changes in this setting. SDFT's stable-fact F1 falls from $\approx$$0.30$ at $s_1$ to $\approx$$0.15$ at $s_3$: by trying to interpolate the new slice against its own previous beliefs, the model ends up corrupting facts that were never supposed to
move. SDPO shows the same effect at smaller magnitude. In contrast, GRPO shows strong performance in learning how to change while preserving performance averaging above baseline across the chain $+1.6$ points.  GRPO witnesses the sharpest jump at $s_2$ of $+6.2$ points which is the largest measured gain of any weight update method.



\noindent \textbf{Prompt evolution shows strong optimization ability on adapting to new knowledge at the cost of forgetting stable knowledge.}   
GEPA reaches the highest drift F1 in the comparison (0.18 on $s_1$, 0.24
on $s_3$) by writing the new fact directly into the playbook in a single
optimization step. The optimizer is rewarded on the validation set,
which contains drift facts; it is not rewarded on stable facts, which
it never sees. So GEPA edits the playbook freely, and the same edits
overwrite stable entries that depend on the same playbook tokens. Stable
F1 falls from 0.32 at $s_1$ to 0.26 at $s_3$.  ACE shows the same
profile although at a smaller magnitude.

\noindent \textbf{Context compression barely influences the base model.} 
For cartridges and In-place TTT drift F1 stays within a few points of baseline across all three slices, with no separation in either direction. Cartridges averages $-0.8$ on drift, with one $+3.2$ inflection on $s_2$ and despite performing better than SDFT and SDPO, but shows no improvement compared to baseline, even harming the performance, belonging to the same category at times.








\subsection{Noisy Temporal Drift}
\label{sec:exp_finance}

From the previous experiments, we observe over longer time horizons, performance steadily degrades. 
In our following temporal drift experiments, we dig deeper into the problem by setting up a task where the model can only improve if it manages to accumulate weak and noisy signals through time.  


\noindent \textbf{Setup.} We construct a challenging temporal data task by taking a realistic case study in finance: predicting whether a stock moves up or down in the 30 days after its 10-K filing. This setting forces a model to operate inside a two-sided time drift. Some patterns in how firms describe risk, growth, and capital plans should be retained, while other signals are regime-specific (e.g. before and after Covid) and stop working as markets evolve. \textit{Doing well now does not guarantee doing well in the future.} A model that loses past knowledge cannot tell whether a new filing is unusual relative to history, and a model that holds too tightly to past years may overfit on spurious or transient regimes.  

We use full-length SEC EDGAR filings (typically 10--13K tokens), with each filing labeled \texttt{up} or \texttt{down} based on its 30-day forward return \citep{lefteris_loukas_2021_5528490, magner2025decoding}. Additional details on data can be found in Appendix~\ref{appendix:temporal}. We train sequentially on filing years 2015 through 2020 in chronological order. After each year, we evaluate on every year in the range to measure both backward transfer (did the model actually learn during training) and forward transfer (did it learn something general it can apply to future years). The signal in any individual filing is known to be weak, but in aggregate a small, tradeable edge can be accumulated\cite{magner2025decoding} (see also Figure \ref{fig:10k_positive_pnl}). At baseline before training, Qwen3-8B sits around 50\% and accuracy is noisy across years. Further details about the experimental setup including relevant hyperparameters are included in Appendix \ref{appendix:temporal}

\begin{figure}[!h]
    \centering
    \includegraphics[width=1\linewidth]{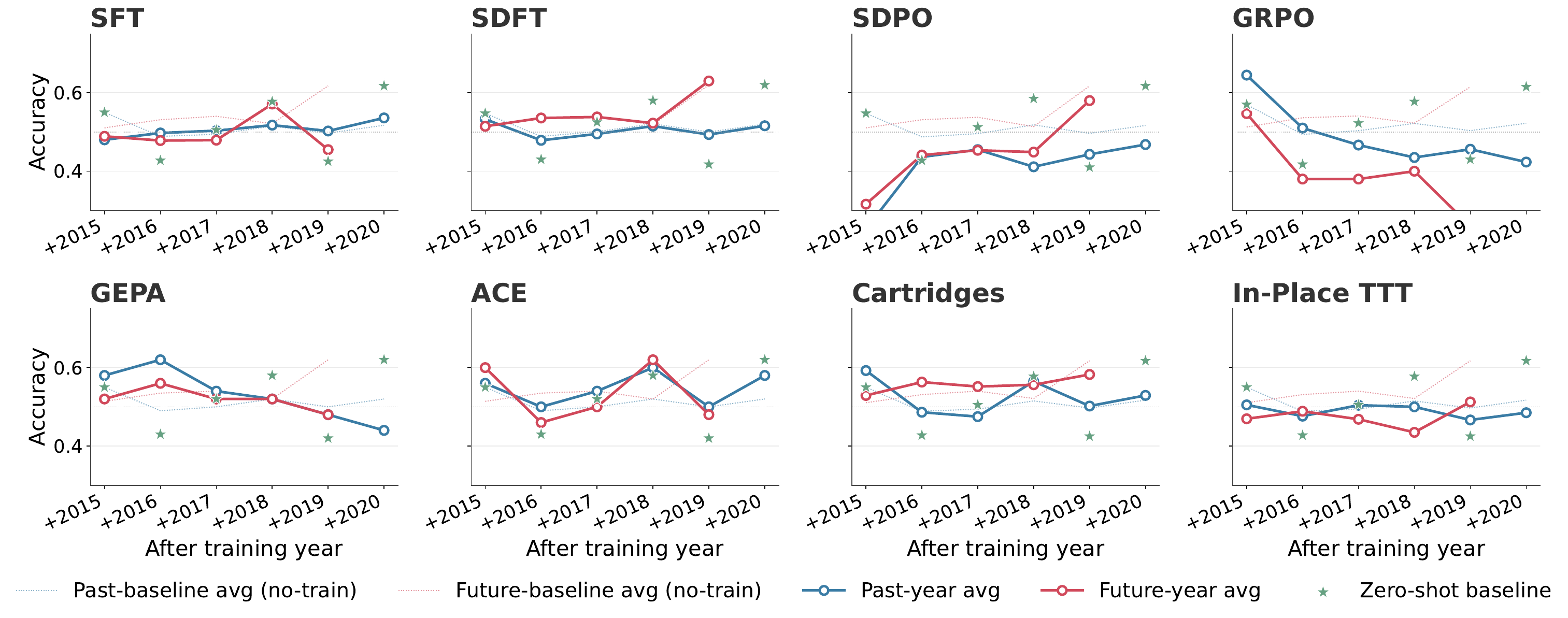}
    \vspace{-2em}
    \caption{Backward and forward accuracy on the 10-K sentiment task, evaluated after sequential training on each year from 2015 to 2020 (Qwen3-8B, no thinking). Blue: average accuracy on past years ($\leq$ current). Red: average accuracy on future years ($>$ current). Stars: zero-shot baseline per year. Distillation-based methods (SDFT, SDPO, Cartridges) sustain forward accuracy, while prompt-based methods (GEPA, ACE) and GRPO degrade on future years. SFT stays near chance throughout.}
    \label{fig:placeholder}
\end{figure}

\noindent \textbf{RL struggles when the reward signal is weak and noisy.} GRPO shows a significant accuracy drop both on forward- and backward-looking sentiment; its success depends heavily on the reward ranking inside each group. If the reward is noisy, the algorithm may reinforce the wrong answer.

\noindent \textbf{Stability-anchored methods show the strongest forward transfer with minimal past degradation.} SDFT's past-year accuracy stays flat around $0.50$ while its future-year accuracy climbs to $\approx$$0.62$ by 2020: distilling against its own previous state avoids both drift and over-fitting to the current year. Cartridges is the most stable method we tested, with past and future hovering around $0.55$--$0.58$ throughout the chain. In-place TTT is the locality reference: its accuracy stays in a tight
34.5--51.5\% band across the chain (mean 46.9\%) and never separates
from baseline. The reset-per-input update geometry is incompatible with
absorbing a slow-moving trend, but it pays no price for trying -- the
method neither acquires nor degrades.

\noindent \textbf{Prompting methods fail to extract useful knowledge for forward transfer.} On this task, similar to RL, GEPA also struggles with learning meaningful information from noisy signals. ACE fits better on past-year data, but also completely fails to predict future years' trend, which suggests that it does not accumulate meaningful knowledge. 



\begin{figure}[!h]
    \centering
    \includegraphics[width=1\linewidth]{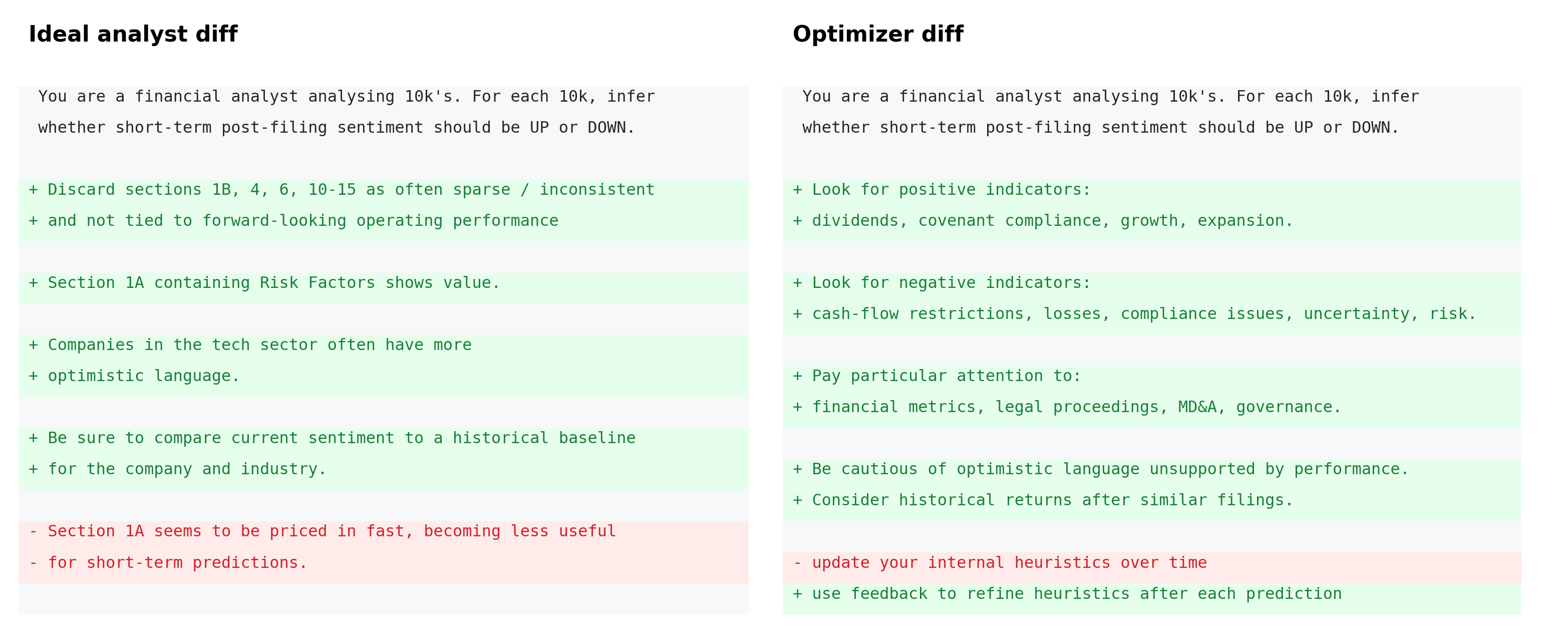}
    \vspace{-2em}
    \caption{Prompt evolution on the 10-K sentiment task: (L) how an idealised analyst may evolve its instructions over time, both noting persistent trends and updating outdated information; (R) the evolution of GEPA shows the addition of generic, static-in-time financial heuristics.}
    \vspace{-1em}
    \label{fig:gepa_prompt_diff}
\end{figure}

Qualitatively, GEPA discovers \emph{generic} financial-analysis heuristics (Figure \ref{fig:gepa_prompt_diff}). These edits fail to capture characteristics of an ideal adaptive analyst e.g. learn to discard irrelevant sections, normalize sentiment relative to sector or company baselines, or distinguish between newly informative elements and those that are already priced in by the market. While it is possible to construct a strategy with a positive Profit and Loss (PnL) with a rule as simple as focusing on relevant sections in the 10-K (see Figure \ref{fig:10k_positive_pnl}), consistent with the observed drop in forward accuracy after optimization, the optimized prompt reduces even this forward PnL. We conclude that in this noisy, time-changing regime, GEPA overfits to transient textual heuristics while failing to preserve exposure to economically meaningful signals that remain predictive out-of-sample.

\subsection{Agentic Task}

The previous two experiments test the time axis under drift that comes from outside the model. Here we test a third sub-regime of time introduced in \S\ref{sec:framework}: the task family stays fixed (actions inside a web app), but the environment's state drifts only because the agent itself acted on it, so state left behind by one step becomes part of the task the model faces at the next.

\noindent \textbf{Data.} We build the agentic suite on top of WebArena-Infinity~\citep{zhou2026wainf}, seeding chain generation from each app's description (which enumerates the available features) and its underlying state files (which fix the entities those features can act on). From these we generate sequential chains in which step~$i{+}1$ depends on a concrete piece of state that step~$i$ leaves behind in the environment: in Gmail, create a label, rename it, then apply the renamed label to a specific email. Each step is paired with a programmatic verifier that reads the app's state directly, so chain success is checked against the environment rather than the agent's narration. Because the sequence arises from the agent's own actions rather than an ordering we impose, we treat it as the most realistic instantiation of continual learning over time.

\noindent \textbf{Setup.} Beyond zero-shot evaluation, we ask whether experience \emph{compounds} across a chain: whether learning from earlier trajectories improves later chain success. We add learning two ways on top of a \texttt{browser-use} agent driving a headless Chromium against the live app. First, an ACE playbook~\citep{zhang2025agentic} on Qwen-32B, which accumulates reusable procedural guidance distilled from prior rollouts. Second, supervised fine-tuning of Qwen-8B on successful traces from a stronger teacher agent, distilling the larger agent's behavior into the smaller model (per-level teachers are given in Appendix~\ref{appendix:agentic}). Within a chain we keep a single agent instance alive across all steps: the browser session is not reset, and each follow-up instruction is appended to the same conversation rather than starting a fresh context, so the model sees its own prior trajectory when acting on step~$i{+}1$. Each step is given a fixed action budget and is scored independently by its verifier. Further details about the experimental setup including relevant hyperparameters are included in Appendix~\ref{appendix:agentic}.

\begin{figure*}[t]
  \centering
  \includegraphics[width=0.8\linewidth]{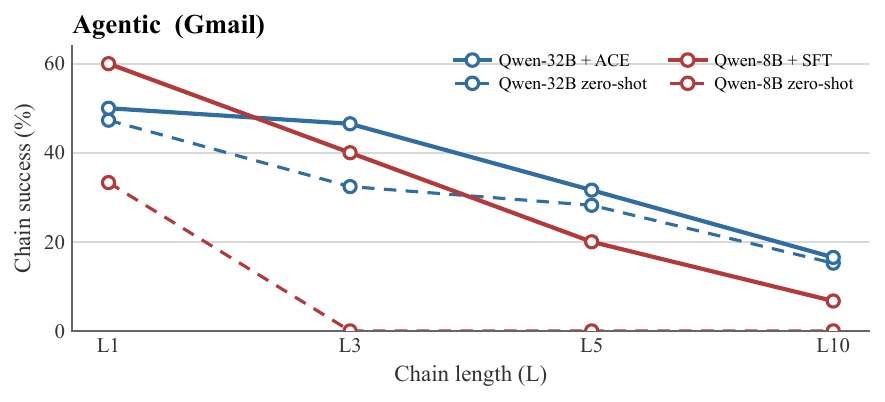}
  \caption{End-to-end chain success rate across chain lengths L1--L10 on the Gmail agentic suite. Color marks the model (blue Qwen-32B, red Qwen-8B); solid lines are the update mechanisms (an ACE playbook on Qwen-32B, and SFT on Qwen-8B distilled from teacher traces) and dashed lines are the corresponding zero-shot baselines. Both mechanisms improve over their baseline at every length, most on the three-step (L3) chains.}
  \label{fig:agentic}
\end{figure*}

\noindent \textbf{Results.} Both update mechanisms raise end-to-end chain success over their no-learning baseline at every horizon length (Figure~\ref{fig:agentic}). The ACE playbook lifts Qwen-32B over its zero-shot baseline at every length, most on the three-step (L3) chains ($32.4\%$ to $46.5\%$). Without training, Qwen-8B collapses to zero past L1; supervised fine-tuning on the teacher's traces recovers substantial success ($40.0\%$ at L3, $20.0\%$ at L5, and $6.7\%$ even at L10), and overtakes both Qwen-32B lines at L1 ($60.0\%$). Absolute success still decays with chain length for every configuration, as accumulated trajectory state grows and later steps must be grounded in increasingly distant environmental cues, converging near the floor by L10. This is the closest setting to deployed use, since the chain comes from the agent's own actions rather than an ordering we picked, and future work could go deeper on this realistic agentic setting.

\section{Discussion}
\subsection{Conclusion} We introduced a unified framework for evaluating continual learning in large language models, defining it as the problem of increasing competence as the world changes along two axes: space (domain shift) and time (temporal drift). By recasting widely used LLM benchmarks as sequential problems and evaluating eight methods under a common mechanism-agnostic protocol, we surface a consistent set of tradeoffs that span across method families. No single method handles all task regimes well: prompt-based methods fit quickly but overwrite prior capabilities; distillation-based methods accumulate knowledge stably but resist new updates; context compression improves efficiency without improving task acquisition; and online reinforcement learning adapts most effectively to factual change but degrades under noisy reward signals. 
Overall, our results suggest that continual learning is not a single capability: different patterns of environmental change require fundamentally different update behaviors. We hope that understanding where each method succeeds and fails will help guide the design of more capable continual learning systems.

\subsection{Limitations and Challenges}
To draw unified conclusions, we use a single model across all evaluations, Qwen3-8B; the relative behavior of the tested methods may change for larger models or models in reasoning mode. In addition, our benchmark suite captures only a subset of realistic environmental change: domain shifts, agentic sequences, noisy temporal drift, and discrete factual updates. While this is a step beyond fixed benchmarks and already reveals that different regimes require different update behaviors, deployed LLMs will face many more open-ended, changing, and uncertain environments and diverse use cases. The results should therefore be read as evidence about which update behaviors are required in different regimes, not as a definitive ranking of continual-learning algorithms.

\section*{Acknowledgement}

We thank Matei Zaharia and Alyosha Efros for helpful discussions and feedback. This work is partially supported by the Laude Institute, Modal, and the ONR MURI N00014-21-1-2801. AH was supported by the National Defense Science and Engineering Graduate (NDSEG) Fellowship Program. This material is based upon work supported by the Air Force Office of Scientific Research under award number FA9550-25-C-B010 in the amount of \$31,607.75.

\bibliographystyle{plainnat}
\bibliography{refs}

\newpage
\appendix
\section{Additional Experimental Details}

\subsection{Compute and Hyperparameters}
\label{app:compute}
 
Within each benchmark, every weight-update method (SFT, SDFT, SDPO,
GRPO) uses the same global batch (32), one epoch over 500 examples
per phase ($\approx$16 optimizer steps), and full-weight updates. The
outer learning rate is set per-method-per-benchmark to keep
cumulative parameter displacement under the empirical collapse
threshold for that label regime~\cite{dohare2024loss}. Each training phase took no longer than 1 hour on a node of 8 H100s.
 
\begin{table}[h]
\centering
\small

\begin{tabular}{@{}llll@{}}
\toprule
\textbf{Method} & \textbf{Domain (B)} & \textbf{Finance (F)} & \textbf{TempWiki (T)} \\
\midrule
SFT  & $1\!\times\!10^{-6}$ & $1\!\times\!10^{-6}$ & $1\!\times\!10^{-6}$ \\
SDFT & $1\!\times\!10^{-5}$ & $1\!\times\!10^{-5}$ & $3\!\times\!10^{-6}$ \\
SDPO & $1\!\times\!10^{-5}$, warmup 10 & $1\!\times\!10^{-5}$, warmup 10 & $3\!\times\!10^{-6}$, warmup 10 \\
GRPO & $1\!\times\!10^{-5}$, warmup 10 & $3\!\times\!10^{-6}$, warmup 10 & $3\!\times\!10^{-6}$, warmup 10 \\
\bottomrule
\end{tabular}
\vspace{1em}
\caption{Outer learning rates for verl-trained methods across
benchmarks. SFT trains on bare-answer gold ($\sim$5--30 tokens), so
per-token gradient signal is roughly $10\times$ the SDFT/SDPO signal,
which trains on $\sim$200+ token rollout golds; lr=$1\!\times\!10^{-5}$
over-displaces the SFT chain in roughly one phase and the model
collapses. The single-token-gold regimes (TempWiki everywhere; finance
for GRPO specifically) collapse to majority-class at
$1\!\times\!10^{-5}$, so we drop those to $3\!\times\!10^{-6}$.}
\label{tab:lr}
\end{table}
 
Cartridges and In-place TTT use their own optimization recipes;
hyperparameters are reported per benchmark in the sections below.

For prompting-based methods, we used cloud API models via OpenRouter. The total cost of all experiments (including development and hyperparamter tuning) was \$1000 USD.

\subsection{Task Settings}
\label{appendix:tasks}

\subsubsection{Domain shift}
\label{appendix:domain}
The learner specialises, in sequence, across three substantively different
domains: tool use (ToolUse~\citep{tang2023toolalpaca}), finance (FinQA~\citep{chen2021finqa}), and
biology (SciKE-Bio~\citep{feng2024sciknoweval}). Each stage provides 500 training instances and a 50-row
held-out evaluation set per task; stages are presented in the canonical
order \texttt{[ToolUse, FinQA, Bio]} so that per-method results are directly
comparable, with permutation-averaged variants reported in the appendix to
control for order effects.

\noindent \textbf{Tasks and data.}
Each phase trains on a 500-example subsample of one task's training
set and validates against all three tasks at every step. The FinQA
test set is unusually large (1{,}125 prompts); we deterministically
subsample it to 128 prompts using the SDPO baseline's seed
(\texttt{np.random.RandomState(1042).permutation(1125)[:128]}, sorted),
so every method evaluates on identical FinQA prompts.
 
\begin{table}[h]
\centering
\footnotesize
\caption{Domain shift benchmark per-phase data. The three tasks
deliberately share no surface-level structure, so a method that
nominally improves task acquisition can still fail by drifting the
model's output format or by overwriting earlier formats once it
learns a later one.}
\label{tab:domain-data}
\begin{tabular}{@{}clrrll@{}}
\toprule
\textbf{Phase} & \textbf{Task} & \textbf{Train} & \textbf{Val} & \textbf{Output format} & \textbf{Gold example} \\
\midrule
1 & Tool Use   & 500 & 50  & \texttt{Action: <tool>...} & \texttt{[\{"Action": "...", "Action\_Input": "\{\}"\}]} \\
2 & FinQA      & 500 & 128 & \texttt{\textbackslash boxed\{<n>\}} & \texttt{\textbackslash boxed\{94.0\}} \\
3 & SciKE-Bio  & 500 & 68  & \texttt{<answer>X</answer>} & \texttt{<answer>B</answer>} \\
\bottomrule
\end{tabular}
\vspace{1em}

\end{table}
 
\noindent \textbf{Per-method hyperparameters.}
SFT, SDFT, SDPO, and GRPO share the optimization budget in
Table~\ref{tab:lr}. SDFT, SDPO, and GRPO additionally use 8 on-policy
rollouts per training prompt (temperature 0.6, top-$p$ 0.95). SDPO
optimizes a DPO-style pairwise objective over scored rollouts (chosen
= high-score, rejected = low-score) against a frozen reference model.
GRPO uses PPO-style clipped objective with per-prompt $z$-scored
advantages over the 8-rollout group.
 
\noindent \textbf{Cartridges configuration.}
Each phase synthesizes a self-study corpus from the cumulative train
pool of phases 1..$i$: Qwen3-8B (running locally as a synthesis
server) is prompted with the four standard seed prompts
(\texttt{use\_case}, \texttt{question}, \texttt{summarization},
\texttt{structuring}) to emit 8{,}192 conversation-style Q/A pairs
grounded in the corpus. Those 8{,}192 conversations are then
context-distilled into a fresh trainable KV cache. Base weights are
frozen. At evaluation the cartridge KV is loaded as a fixed prefix
and the validation prompt is generated against it.
 
\begin{table}[h]
\caption{Cartridges configuration on the domain shift benchmark.}
\centering
\small
\begin{tabular}{@{}ll@{}}
\toprule
\textbf{Param} & \textbf{Value} \\
\midrule
Synthesizer & Qwen3-8B (local server) \\
Synthesis seed prompts & \texttt{use\_case}, \texttt{question}, \texttt{summarization}, \texttt{structuring} \\
Synthesized Q/A pairs & 8{,}192 / phase \\
Trainable KV length & 2{,}048 tokens \\
KV initialization & \texttt{KVFromRandomText} (no carry-over between phases) \\
Distillation epochs & 1 \\
Global batch & 32 \\
Learning rate & $2\!\times\!10^{-2}$ \\
Trainable params & KV cache only \\
\bottomrule
\end{tabular}
\vspace{1em}

\end{table}

\paragraph{In-place TTT configuration.}
Continual pretraining via VeOmni plaintext-iterable JSONL with one
record per training prompt (cumulative pool of phases 1..$i$ in
order B). Fast-weight updates are activated on a sparse layer set
(paper-recommended for continual training). The chain hands each
phase's HF-format checkpoint forward as the next phase's
\texttt{model\_path}.

\begin{table}[h]
\caption{In-place TTT configuration on the domain shift benchmark. The
\texttt{kl} target is for from-scratch context lengthening and is not
used here.}
\centering
\small
\begin{tabular}{@{}ll@{}}
\toprule
\textbf{Param} & \textbf{Value} \\
\midrule
Outer-loop steps / phase & 50 \\
Global batch & 64 \\
Micro-batch & 1 \\
Sequence length & 16{,}384 \\
Outer optimizer & AdamW \\
Outer-loop LR & $5\!\times\!10^{-6}$, cosine decay (warmup 0.02, decay ratio 0.90) \\
Weight decay & 0.1 \\
Max grad norm & 1.0 \\
Attention backend & \texttt{flash\_attention\_2} \\
\texttt{ttt\_layers} & \texttt{[0, 6, 12, 18, 24, 30, 36]} \\
\texttt{ttt\_mode} / \texttt{ttt\_proj} & \texttt{True} / \texttt{True} \\
\texttt{ttt\_lr} (inner-loop fast-weight LR) & 3 \\
\texttt{ttt\_chunk} & 4096 \\
\texttt{ttt\_target} & \texttt{hidden\_states} (continual-training mode) \\
Sharding & FSDP2, mixed precision, gradient checkpointing \\
Source corpus & Cumulative training prompts of phases 1..$i$ in order B \\
\bottomrule
\end{tabular}
\vspace{1em}

\end{table}

\noindent \textbf{Scoring.}
Each task has its own scorer. A prediction in the
wrong format scores zero --- the benchmark measures task knowledge
\emph{and} format-fidelity together.
\begin{itemize}
\item \textbf{Tool-use:} regex-extract every \texttt{Action: <name>}
  and \texttt{Action Input: \{...\}} line; parse each
  \texttt{Action\_Input} body as JSON; accept iff the multiset of
  predicted action names equals the gold and the merged
  \texttt{Action\_Input} dict equals the gold dict key-by-key.
\item \textbf{FinQA:} extract the contents of the last
  \texttt{\textbackslash boxed\{...\}}; accept iff
  $|\text{pred} - \text{gold}| \le \max(0.005 \cdot |\text{gold}|, 0.01)$
  (FinQA paper convention; exact-match would penalize harmless
  rounding).
\item \textbf{SciKE-Bio:} look for a strict \texttt{<answer>X</answer>}
  tag; accept iff the single-letter \texttt{X} equals the gold letter.
\end{itemize}
 
 \newpage
\subsubsection{TempWiki Benchmark Details}
\label{appendix:wiki}
\noindent \textbf{Task and data.}
Wikipedia is mirrored at four monthly snapshots --- 2025-11-20,
2025-12-01, 2026-01-01, 2026-02-01 --- and for each adjacent pair we
extract the drift set of (subject, relation, object) triples whose
object value changed across snapshots. We also extract a stable set
of (subject, relation) pairs whose answer did not change across the
chain, and we hold this set out across all phases as a forgetting
probe.

Using these snapshots, we construct a chain with three phases corresponding to three diff-pair slices, with 500 training drift triples, 50 validation drift triples, and
50 stable validation triples. Existing temporal-knowledge benchmarks such as the original TemporalWiki~\cite{jang2022temporalwiki} stop at facts that have changed and do not
audit unchanged knowledge, which makes the catastrophic-memorization
failure mode invisible. Pairing the drift set with a held-out stable
probe is what lets the matrix expose the trade-off methods make between
acquiring the new fact and preserving the surrounding knowledge.
 
\begin{table}[h]
\caption{TempWiki per-phase data. The validation prompt is
intentionally terse: a system message instructing the model to emit
only the object string, a user message of the form
\texttt{<subject> <relation>}, and a gold object string. Two
illustrative examples: (drift) user
``\textit{S\~{a}o Paulo twinned administrative body}'' $\to$ gold
``\textit{Buenos Aires}''; (stable) user
``\textit{Jibra'il Dallal place of death}'' $\to$ gold
``\textit{Aleppo}''.}
\centering
\small
\begin{tabular}{@{}clrrr@{}}
\toprule
\textbf{Phase} & \textbf{Slice (snapshots)} & \textbf{Train drift} & \textbf{Val drift} & \textbf{Stable val (held constant)} \\
\midrule
1 (ts1) & Nov $\to$ Dec 2025      & 500 & 50 & 50 \\
2 (ts2) & Dec 2025 $\to$ Jan 2026 & 500 & 50 & 50 \\
3 (ts3) & Jan $\to$ Feb 2026      & 500 & 50 & 50 \\
\bottomrule
\end{tabular}
\vspace{1em}

\end{table}

\noindent \textbf{Per-method hyperparameters.}
SFT/SDFT/SDPO/GRPO use the recipe from 
the 500 drift triples Q/A-formatted directly; stable triples are held
out and never appear in training. Cartridges uses the same
configuration as on the domain benchmark, with two adjustments: the
source corpus is the cumulative Wikipedia article bodies rather than
the cumulative training pool, and the slice-1 article pool
deliberately includes the stable-val subjects so the synthesizer has
reason to write Q/A about them; only drift articles are added on
later slices, which keeps the stable probe legitimately held out from
the gradient-equivalent updates. In-place TTT uses the same
fast-weight recipe as on the 10-K benchmark; per-article records are
formatted as \texttt{=== Wikipedia article  title=<sitelink>
as\_of=<snapshot> ===\textbackslash n<body>\textbackslash n} followed
by an optional \texttt{Known facts:} block holding the canonical
\texttt{subject relation -> gold} lines.

\begin{table}[h]
\caption{Cartridges configuration on the TempWiki benchmark.}
\centering
\small
\begin{tabular}{@{}ll@{}}
\toprule
\textbf{Param} & \textbf{Value} \\
\midrule
Synthesizer & Qwen3-8B (local server) \\
Synthesis seed prompts & \texttt{use\_case}, \texttt{question}, \texttt{summarization}, \texttt{structuring} \\
\texttt{min\_tokens\_per\_chunk} / \texttt{max\_tokens\_per\_chunk} & 1{,}024 / 2{,}048 \\
Synthesized Q/A pairs & 8{,}192 / phase \\
Trainable KV length & 2{,}048 tokens \\
KV initialization & \texttt{KVFromRandomText} (no carry-over between phases) \\
\texttt{top\_k\_logits} (distillation) & 20 \\
\texttt{packed\_seq\_length} / \texttt{packing\_mode} & 2{,}048 / \texttt{truncate} \\
Distillation epochs & 1 \\
Global batch & 32 \\
Learning rate & $2\!\times\!10^{-2}$ \\
Source corpus & Cumulative Wikipedia article bodies (slices 1..$i$) \\
Articles / slice & $\sim$528 (cumulative $\approx$1{,}484 after phase 3) \\
Trainable params & KV cache only \\
\bottomrule
\end{tabular}
\vspace{1em}

\end{table}

\begin{table}[h]
\caption{In-place TTT configuration on the TempWiki benchmark.}
\centering
\small
\begin{tabular}{@{}ll@{}}
\toprule
\textbf{Param} & \textbf{Value} \\
\midrule
Outer-loop steps / phase & 50 \\
Global batch & 64 \\
Micro-batch & 1 \\
Sequence length & 16{,}384 \\
Outer optimizer & AdamW \\
Outer-loop LR & $5\!\times\!10^{-6}$, cosine decay (warmup 0.02, decay ratio 0.90) \\
Weight decay & 0.1 \\
Max grad norm & 1.0 \\
Attention backend & \texttt{flash\_attention\_2} \\
\texttt{ttt\_layers} & \texttt{[0, 6, 12, 18, 24, 30, 36]} \\
\texttt{ttt\_mode} / \texttt{ttt\_proj} & \texttt{True} / \texttt{True} \\
\texttt{ttt\_lr} (inner-loop fast-weight LR) & 3 \\
\texttt{ttt\_chunk} & 4096 \\
\texttt{ttt\_target} & \texttt{hidden\_states} (continual-training mode) \\
Sharding & FSDP2, mixed precision, gradient checkpointing \\
Source corpus & Cumulative Wikipedia article bodies (slices 1..$i$) \\
Per-article header & \texttt{=== Wikipedia article  title=<sitelink>  as\_of=<snapshot> ===} \\
\bottomrule
\end{tabular}
\vspace{1em}

\end{table}

\noindent \textbf{Scoring (word-level F1, paraphrase-permissive).}
The scorer computes word-level F1. The prediction is stripped of
\texttt{<think>} traces and chat suffixes; the model's answer is taken
to be the contents of an explicit \texttt{Answer: ...} line if
present, otherwise the first non-empty line. Normalization (applied
identically to prediction and gold): lower-case, strip the articles
\texttt{a}, \texttt{an}, \texttt{the}, strip all ASCII punctuation,
collapse whitespace, tokenize by whitespace into a token bag.
 
We score predictions with a token-level F1 against the gold answer.
Both strings are normalized (lowercased, articles and punctuation stripped,
whitespace collapsed) and split into tokens. Let $S$ be the number of
tokens shared between the prediction and the gold (counted with
repetition). Then
\[
P = \frac{S}{N_{\text{pred}}}, \qquad
R = \frac{S}{N_{\text{gold}}}, \qquad
F_1 = \frac{2PR}{P + R}.
\]
A prediction counts as a hit if $F_1 \ge 0.5$, and zero otherwise.
We report \textsc{score\_mean@8}: the average hit rate across 8 rollouts
per prompt and across all prompts in a slice.

\subsubsection{TempWiki-Easy: a filtered, continuously-scored variant}
\label{appendix:wiki_easy}

The TempWiki drift set in the main text (Fig.~\ref{fig:tempwiki})
pools \emph{every} relation type whose object value changed between two
snapshots. Two properties of that construction make the
catastrophic-memorizing signal noisier than it should be. First, many
Wikipedia diffs are not genuine world-knowledge updates: reformatting,
disambiguation edits, list re-orderings, and annotation-style relations
change the object string without there being a well-posed ``new fact'' to
learn. Second, the object-F1 scorer used there is effectively binary at
the string level, so a near-correct object (e.g.\ a partial or alternately
formatted name) is scored identically to a wholly wrong one. Both effects
inflate the apparent divergence between drift and stable curves. TempWiki-Easy
is a cleaned re-cut of the same benchmark designed to isolate the update
behavior from these artifacts.

\noindent \textbf{Construction.}
We keep the four monthly snapshots and the identical three-slice chain as
TempWiki (Nov$\to$Dec 2025, Dec 2025$\to$Jan 2026, Jan$\to$Feb 2026), and
apply three changes to the diff extraction:
\begin{itemize}[leftmargin=1.4em,itemsep=1pt,topsep=1pt]
  \item \textbf{Category allow-list.} Only (subject, relation) pairs drawn
  from an allow-list of relation types for which ``change'' is
  well-defined (single-valued, factual, verifiable object) are eligible.
  Relations prone to formatting or annotation churn are excluded outright,
  which removes the pathological categories responsible for spurious
  drift/stable divergence in the original cut.
  \item \textbf{Aggressive judged filtering.} Each surviving candidate diff
  is passed through an LLM judge that keeps the triple only if the edit
  reflects a semantically meaningful factual update between the two
  snapshots, discarding cosmetic or ambiguous edits.
  \item \textbf{QA regeneration + continuous scoring.} We regenerate
  question/answer pairs over the filtered facts and, rather than
  thresholding the word-level F1 into a hit at $F_1\!\ge\!0.5$ (the
  \textsc{score\_mean@8} hit rate used in Appendix~\ref{appendix:wiki}),
  report the continuous word-level $F_1$ directly. Partially correct
  objects receive partial credit, so small real movements in knowledge
  are visible rather than being quantized to $0/1$.
\end{itemize}

\begin{figure}[!h]
    \centering
    \includegraphics[width=\linewidth]%
        {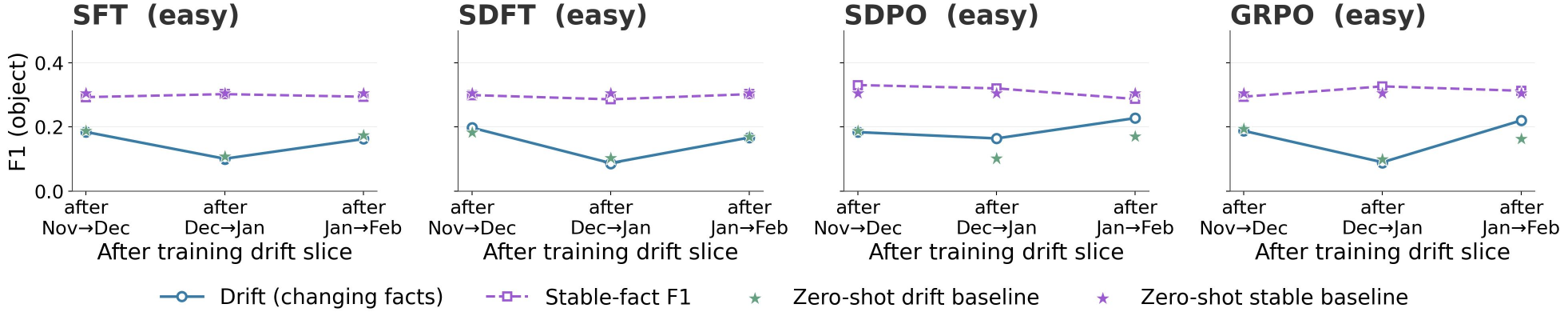}
    \vspace{-1em}
    \caption{TempWiki-Easy: object F1 across the three drift slices
    (Qwen3-8B, no thinking, seed 42). }
    \label{fig:twiki_easy}
\end{figure}

The most visible difference from TempWiki Fig.~\ref{fig:tempwiki} is that the
stable-fact trajectory (purple dashed) no longer trades off against drift.
In the original cut, SDFT's stable F1 fell from  to
and SDPO's collapsed to  by the final slice; in
TempWiki-Easy both remain flat and essentially coincident with the
zero-shot stable baseline across all three slices.  With the confound removed, most methods sit on top of the zero-shot drift
baseline: SFT and SDFT neither learn the changed facts nor harm the stable
ones. The
two exceptions are the reinforcement/distillation updaters. SDPO climbs to
0.22 on the final slice, above its 0.17 zero-shot
baseline, while its stable F1 drifts only mildly 
GRPO shows the largest net effect: after the same mid-chain dip it reaches
$\sim$0.22 drift on Jan$\to$Feb  while holding
stable F1 flat.

\subsubsection{Temporal Drift}
\label{appendix:temporal}
\noindent \textbf{10-K Sentiment Benchmark Details}

The core task is fixed (predict the directional sentiment of a 10-K
filing) \citep{lefteris_loukas_2021_5528490, magner2025decoding} but slides across calendar time: each stage corresponds to one fiscal
year of 10-K filings, presented in chronological order from 2015 to 2020.
After training on year~$y$ we evaluate on every other year, splitting the
score into \emph{past-year} accuracy ($\le y$) and \emph{future-year}
accuracy ($> y$). The target is intentionally weak-signal: real financial
text is informative only in aggregate, and the question is precisely whether
a learner can stay extract rules to align with a slow-moving, noisy trend rather than rely on spurious local-in-time patterns.

A second temporal benchmark constructed from Wikipedia revisions captured
between November 2025 and February 2026 \citep{jang2022temporalwiki}. Stages correspond to three
chronological slices ($s_1, s_2, s_3$) of facts whose object value changes
over the window. We evaluate object-level F1 on each slice after every stage,
together with F1 on a control set of \emph{stable} (unchanged) facts. This
exposes a failure mode invisible in standard CL benchmarks: methods that
successfully update drifted facts may simultaneously degrade unchanged ones
---the mirror image of forgetting.

\noindent \textbf{Task and data.}

Specifically, we construct a six-year chain of 10-K filings drawn from SEC EDGAR~\cite{lefteris_loukas_2021_5589195, financial-reports-sec},
covering fiscal years 2015 through 2020 in chronological order. Each
phase contains $\sim$500 training filings and 50 held-out validation
filings drawn from that year, with each filing represented in full
(typically 10--13K tokens) wrapped between custom
\texttt{[START OF FILING]}~/~\texttt{[END OF FILING]} markers. Cumulative
training tokens grow from $\sim$6M after phase~1 to $\sim$37M after
phase~6. The label is a single token, \texttt{up} or \texttt{down},
derived from the issuer's 30-day forward simple stock return joined to
the filing by Central Index Key (CIK) and filing date. Prior
work on financial NLP either treats each year as an independent task
or pools all years; we hold the task fixed and let the underlying
data-generating process drift, exposing methods to the actual condition
under which deployed financial models age out of distribution.

In this task,the model predicts the directional sign of an issuer's 30-day forward simple stock return from the textual content of its annual 10-K
filing. The chain has six sequential phases for filing years 2015
through 2020 in chronological order, producing a 6$\times$6 matrix.
Each filing is full-length SEC EDGAR text wrapped between custom
\texttt{[START OF FILING]} / \texttt{[END OF FILING]} markers
(typically 10--13K tokens); the gold label is a single token,
\texttt{up} or \texttt{down}, derived from a precomputed returns
table joined by issuer CIK and filing date.
 
 
\noindent \textbf{Per-method hyperparameters.}
SFT/SDFT/SDPO/GRPO use the optimization recipe from
Table \ref{app:compute} with two finance-specific data settings:
\texttt{data.max\_prompt\_length} = 16{,}384 to fit the 10-K body, and
\texttt{data.max\_response\_length} = 256 since the response is a
single token plus chat tail.
 
\noindent \textbf{Cartridges configuration.}
Per phase, concatenate the cumulative pool of all filings seen so far
(years 1..$i$, shuffled at filing granularity); synthesize 8{,}192
self-study Q/A pairs grounded in that pool; distill into a fresh
2{,}048-token KV cache. The cartridge is loaded as a prefix at
evaluation, with the full filing body still included in the user
message. 
 
\noindent \textbf{In-place TTT configuration.}
Continual pretraining on the same cumulative corpus, written as
VeOmni plaintext-iterable JSONL with one record per filing (full
body). Fast-weight updates are activated on a sparse layer set
(paper-recommended for continual training). The chain hands each
phase's converted HF checkpoint forward as the initial weights for
the next phase.
 
\begin{table}[h]
\caption{In-place TTT configuration on the 10-K benchmark. The
\texttt{kl} target is for from-scratch context lengthening and is not
used here.}
\centering
\small
\begin{tabular}{@{}ll@{}}
\toprule
\textbf{Param} & \textbf{Value} \\
\midrule
Outer-loop steps / phase & 50 \\
Global batch & 64 \\
Micro-batch & 1 \\
Sequence length & 16{,}384 \\
Outer-loop LR & $5\!\times\!10^{-6}$, cosine decay (warmup 0.02, decay ratio 0.90) \\
Weight decay & 0.1 \\
Max grad norm & 1.0 \\
\texttt{ttt\_layers} & \texttt{[0, 6, 12, 18, 24, 30, 36]} \\
\texttt{ttt\_lr} (inner-loop fast-weight LR) & 3 \\
\texttt{ttt\_chunk} & 4096 \\
\texttt{ttt\_target} & \texttt{hidden\_states} (continual-training mode) \\
Sharding & FSDP2, mixed precision, gradient checkpointing \\
Source corpus & Cumulative filings, years 1..$i$ \\
\bottomrule
\end{tabular}
\vspace{1em}
\end{table}
 
\noindent \textbf{Scoring.}
The prediction string is stripped of any \texttt{<think>} trace and
chat suffixes, then matched against the case-insensitive whole-word
patterns \verb|\bup\b| and \verb|\bdown\b|. The first match wins;
ties break in favor of \texttt{up}. A prediction is correct iff the
matched token equals the gold.

\noindent \textbf{Further Details on the 10K sentiment task with GEPA}

\begin{figure}[!h]
    \centering
    \includegraphics[width=\linewidth]{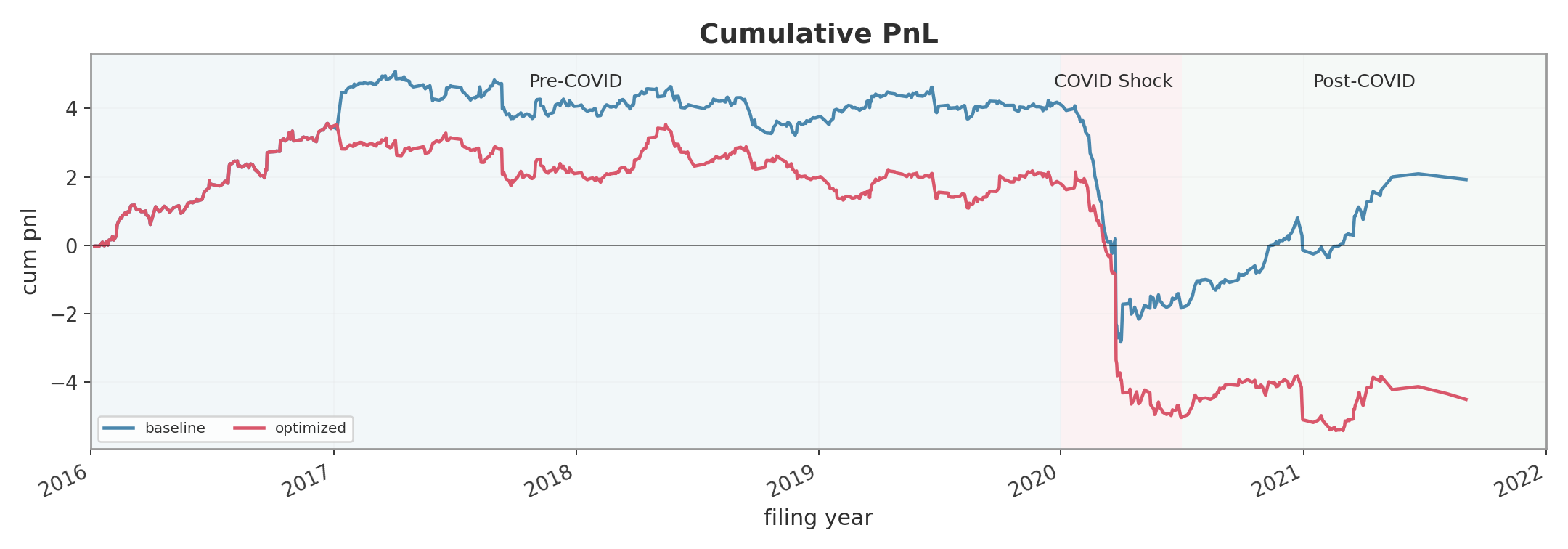}
    \caption{A PnL constructed from retaining only Items 2, 5, 7, 7A, 8, 9, and 9A from the 10-K filings. These sections capture the firm’s physical operating footprint, shareholder and capital-allocation information, management’s discussion of performance and risks, market-risk exposures, core financial statements, auditor changes or disagreements, and internal-control quality.}
    \label{fig:10k_positive_pnl}
\end{figure}

\noindent \textbf{PnL computation.}
We construct a daily long-short strategy from model predictions. For each filing observed on day $t$, the model predicts a directional position $p_i \in {-1, +1}$, corresponding to DOWN or UP sentiment respectively. Let $r_i$ denote the realized 30-day return following the filing. Daily portfolio returns are computed as the average position-weighted return across all filings observed on that day $R_t = \frac{1}{N_t} \sum_{i=1}^{N_t} p_i r_i$,
where $N_t$ is the number of filings on day $t$. The cumulative PnL is obtained by cumulatively summing the daily returns $R_t$ over time. The optimized PnL curve is causal. After optimizing the prompt on year $X$, predictions generated using the optimized prompt are evaluated only on year $X+1$, reflecting a sequential deployment setting. Figure \ref{fig:10k_positive_pnl} shows that it is possible to construct a strategy with a positive PnL pre-Covid through a simple rule: attend only to a subset of sections. Similar to the observed accuracy drop in Figure \ref{fig:placeholder}, GEPA's optimisation fails to generalise in a causal forward-looking manner, dropping PnL performance. 

\begin{figure}[!htbp]
    \centering
    \includegraphics[width=\linewidth]{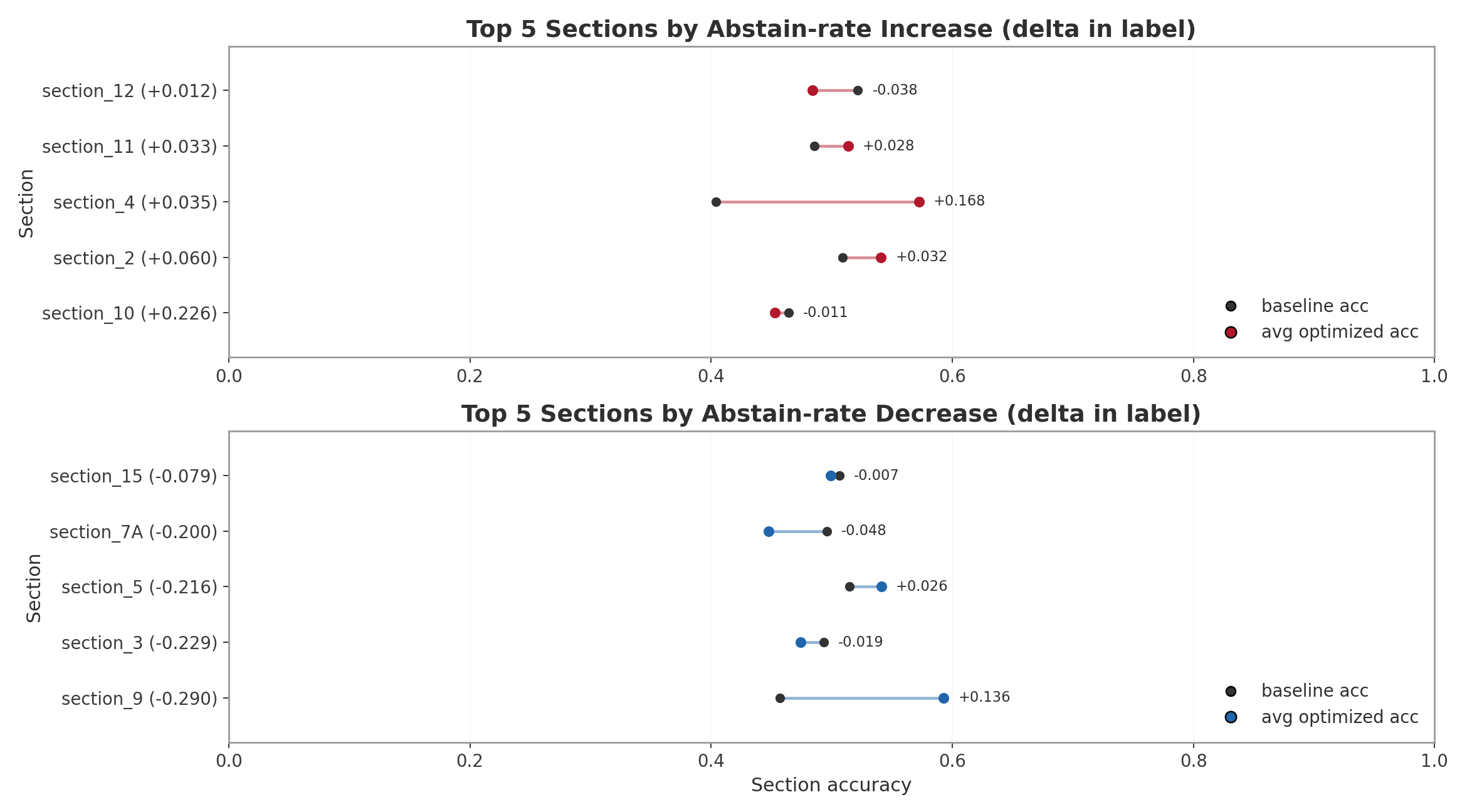}
    \caption{The change in abstain-rate per section. On the aggregate, predicting more often (abstain-rate decrease) or abstaining more often (abstain-rate increase) does not consistently lead to an increase in accuracy, showing that GEPA fails to extract section-level usefulness.}
    \label{fig:10k_sentiment_abstain}
\end{figure}

\noindent \textbf{Section-wise sentiment with possibility to abstain}
We additionally explored a setup in which the model was allowed to abstain from a prediction for a given filing sections during GEPA optimization, as not all sections are considered equally informative. Figure \ref{fig:10k_sentiment_abstain} reports the relative change in section utilization frequency under this setup. Some sections became increasingly downweighted, while others were used more heavily by the optimized prompts. Notably, some down or upweighted sections indeed led to accuracy increases, but not consistently. Equally, Section 1A (“Risk Factors”), which prior work has shown to contain predictive risk sentiment information~\cite{magner2025decoding}, remained largely unchanged. This may indicate that the optimizer failed to infer its predictive usefulness.

\subsubsection{Agentic.}
\label{appendix:agentic}

\paragraph{Update-mechanism hyperparameters.} The two update mechanisms plotted in Figure~\ref{fig:agentic} are configured as follows.

\emph{ACE (Qwen3-32B).} An online batch pipeline runs each batch of $5$ chains live in the browser, then reflects and curates. The reflector runs once per chain, tagging each existing bullet helpful, harmful, or neutral; the curator runs once per batch over the combined batch reflections with ADD-only operations under a $50{,}000$-token budget. The reflector/curator LLM is Bedrock \texttt{qwen.qwen3-32b-v1:0} via the Converse API, with \texttt{temperature}$=0.3$ and \texttt{maxTokens}$=4096$. The playbook has five fixed sections (\texttt{strategies\_and\_insights}, \texttt{common\_mistakes\_to\_avoid}, \texttt{problem\_solving\_heuristics}, \texttt{ui\_navigation\_tips}, \texttt{others}) and bullets in the format \texttt{[slug-NNNNN] helpful=H harmful=X :: content}. We run $1$ epoch over $5$ batches ($\approx$25--30 chains, no held-out test) and $5$ seeds (\texttt{seed0}--\texttt{seed4}), each shuffling the chain order. At evaluation the playbook is prepended to the browser-use system prompt, the agent (qwen3-32b) decodes at factory-default settings, and each cell covers $5$ seeds $\times$ $30$ chains. Per-horizon action budgets and timeouts are L1: $10$ steps / $300$s, L3: $20$ / $900$s, L5: $25$ / $1200$s, L10: $40$ / $1800$s.

\emph{SFT (Qwen3-8B).} We fine-tune Qwen3-8B with LoRA (rank $32$, $\alpha=64$) on the \texttt{q,k,v,o,gate,up,down} projections, learning rate $5\times10^{-5}$, gradient accumulation $4$, maximum sequence length $16{,}384$, for $1$ epoch, since training beyond one epoch overfit and decreased performance on some task-suite tasks. Trajectories are distilled from a teacher agent (Qwen3-32B for L1--L5, Qwen3-235B for L10); per-horizon trajectory and training-example counts are given in Table~\ref{tab:agentic-sft-data}.

\begin{table}[ht!]
  \centering
  \caption{Distillation data for the SFT mechanism, by horizon length. Traj.: kept (successful) trajectories; Train/Val: trajectory split; Steps: training-step examples.}
  \label{tab:agentic-sft-data}
  \small
  \setlength{\tabcolsep}{4pt}
  \begin{tabular}{lrrrrrrl}
    \toprule
    Level & Tasks & Traj. & Traj./task & Train/Val & Steps & Steps/traj & Teacher \\
    \midrule
    L1  &  46 &   612 & 13.3 &   551 / 61  &  5{,}485 & $\sim$10 & Qwen3-32B \\
    L3  &  96 & 1{,}360 & 14.2 & 1{,}224 / 136 & 22{,}074 & $\sim$18 & Qwen3-32B \\
    L5  & 132 & 1{,}536 & 11.6 & 1{,}383 / 153 & 32{,}199 & $\sim$23 & Qwen3-32B \\
    L10 & 242 & 2{,}299 &  9.5 & 2{,}070 / 229 & 69{,}674 & $\sim$34 & Qwen3-235B \\
    \bottomrule
  \end{tabular}
\end{table}

\paragraph{Suite construction.} We build the agentic suite on top of WebArena-Infinity~\citep{zhou2026wainf}. Chain generation is seeded from two sources per app: the app's description, which enumerates the available features, and its underlying state files, which fix the concrete entities those features can act on. Conditioning on both lets us generate chains whose steps reference entities that actually exist in the environment. From these we generate sequential chains of length $1$, $3$, $5$, and $10$, in which step~$i{+}1$ depends on a concrete piece of state that step~$i$ leaves behind in the environment. In Gmail, for instance, the agent creates a label, renames it, and then applies the renamed label to a specific email, so that each later step is only well-defined given the state produced by the earlier ones. The full suite contains $510$ chains across four apps (Gmail, Gmail Contacts, Handshake, PayPal), spanning the four horizon lengths.

\paragraph{Verification.} Each step is paired with a programmatic verifier that reads the app's state directly rather than parsing the agent's narration, so success is checked against the environment. A step is marked correct only when its verifier confirms the intended state change, and a chain of length $L$ succeeds only if all $L$ of its steps pass. Grounding verification in state ensures that an agent receives no credit for claiming to have completed a step it did not actually perform.

\paragraph{Agent harness.} All runs use a \texttt{browser-use} agent driving a headless Chromium against the live app. Within a chain we keep a single agent instance alive across all steps: the browser session is not reset between steps, and each follow-up instruction is appended to the same conversation rather than starting a fresh context, so that when acting on step~$i{+}1$ the model sees its own prior trajectory---observations, actions, and errors---from the preceding steps. Each step is given a fixed action budget and is scored independently by its verifier.

\paragraph{Zero-shot models.} In addition to the learning methods reported in the main body, we evaluate Deepseek-v3.2~\citep{liu2025deepseek}, GPT-OSS-120B~\citep{agarwal2025gpt}, and Gemini-2.5-flash-lite~\citep{comanici2025gemini} zero-shot under the identical harness. All three degrade sharply as chain length grows (Figure~\ref{fig:agentic}). The decay tracks the point at which the accumulated trajectory approaches the model's context window: once earlier trajectory state has to be summarized to fit, the model loses access to the precise environmental cues it needs to ground later actions. Gemini's larger context window does not rescue this failure mode---it degrades faster than Deepseek despite more available context---suggesting that the bottleneck is the agent's ability to actually \emph{use} long trajectory information rather than raw context capacity.

The same ``does experience stick across tasks'' question, but where the
chain arises from the agent's own actions rather than from a fixed
ordering we picked. A learner operates in sequential chains of web tasks
built on WebArena~Infinity \cite{zhou2026wainf}, where the environment produced by task~$i$ is
the starting state of task~$i{+}1$: an action taken early in the chain
(an archived email, an applied label, a navigation state) persists and
affects every subsequent task's success. We measure end-to-end pass rate
as a function of chain length $L \in \{1, 3, 5, 10\}$, across three apps
(Gmail, Gmail~Accounts, PayPal~Wallet) and three backbones (Deepseek-v3.2 ~\cite{liu2025deepseek},
Gemini-2.5-Flash-Lite~\cite{comanici2025gemini}, GPT-OSS-120B~\cite{agarwal2025gpt}). This is the closest of our benchmarks to deployed
agentic use; we treat it as the natural-regime counterpart to Domain
shift's controlled regime.

%

\begin{table}[ht!]
  \centering
  \caption{Chain pass rate on the \texttt{gmail} app, by model and horizon length.
  Denominator is 30 chains at every length.}
  \label{tab:agentic-gmail}
  \begin{tabular}{lcrrrr}
    \toprule
    Model & Context & L1 & L3 & L5 & L10 \\
    \midrule
    deepseek-v3.2          & 128k & 70\% & 30\% & 23\% & 0\% \\
    gemini-2.5-flash-lite  & 1M   & 47\% &  7\% &  7\% & 0\% \\
    gpt-oss-120b           & 128k & 60\% & 40\% & 17\% & 0\% \\
    \bottomrule
  \end{tabular}
\end{table}

\begin{table}[ht!]
  \centering
  \caption{Chain pass rate on the \texttt{gmail-contacts} app, by model and horizon length.
  Denominators are 55/32/20/12 chains at L1/L3/L5/L10.}
  \label{tab:agentic-contacts}
  \begin{tabular}{lcrrrr}
    \toprule
    Model & Context & L1 & L3 & L5 & L10 \\
    \midrule
    deepseek-v3.2          & 128k & 75\% & 91\% & 85\% & 8\% \\
    gemini-2.5-flash-lite  & 1M   & 49\% & 25\% &  5\% & 0\% \\
    gpt-oss-120b           & 128k &  40\% &  21\% &  0\% & 0\% \\
    \bottomrule
  \end{tabular}
\end{table}

\begin{table}[ht!]
  \centering
  \caption{Chain pass rate on the \texttt{paypal} app, by model and horizon length.
  Denominators are 50/24/14/12 chains at L1/L3/L5/L10.}
  \label{tab:agentic-paypal}
  \begin{tabular}{lcrrrr}
    \toprule
    Model & Context & L1 & L3 & L5 & L10 \\
    \midrule
    deepseek-v3.2          & 128k & 80\% & 67\% &  7\% & 17\% \\
    gemini-2.5-flash-lite  & 1M   & 72\% & 29\% &  0\% &  0\% \\
    gpt-oss-120b           & 128k & 68\% & 29\% & 14\% &  0\% \\
    \bottomrule
  \end{tabular}
\end{table}

\begin{table}[ht!]
  \centering
  \caption{Chain pass rate on the \texttt{handshake} app, by model and horizon length.
  Denominators are 70/41/30/30 chains at L1/L3/L5/L10.}
  \label{tab:agentic-handshake}
  \begin{tabular}{lcrrrr}
    \toprule
    Model & Context & L1 & L3 & L5 & L10 \\
    \midrule
    deepseek-v3.2          & 128k & 80\% & 37\% & 27\% & 7\% \\
    gemini-2.5-flash-lite  & 1M   & 44\% &  7\% &  0\% & 0\% \\
    gpt-oss-120b           & 128k & 57\% & 17\% &  7\% & 0\% \\
    \bottomrule
  \end{tabular}
\end{table}

 \section{Broader Impact}
 
 This work develops evaluation infrastructure for continual learning in large language models, with the goal of helping the research community build systems that remain competent as the world changes. Better continual learning methods could have positive societal impact by reducing the need to retrain large models from scratch as new information becomes available, lowering the associated computational cost and energy consumption. At the same time, systems that can efficiently update their knowledge after deployment also carry risks: a model that rapidly incorporates new information could more easily be steered toward harmful or false beliefs through targeted data injection. Our work does not propose a new training method, but to the extent that it accelerates progress in continual learning, these dual-use considerations apply. We do not anticipate direct harms from the framework or evaluation protocols introduced here, as they are designed for measuring and understanding model behavior rather than enabling any specific application.

\end{document}